\documentclass[runningheads]{llncs}
\ifdefined\pdfminorversion\pdfminorversion=7\fi

\usepackage[T1]{fontenc}
\usepackage{graphicx}
\usepackage{microtype}
\usepackage{cite}
\usepackage{amsmath,amssymb}
\usepackage{algorithm}
\usepackage{algpseudocode}
\usepackage{booktabs}
\usepackage{array}
\usepackage{tabularx}
\usepackage{placeins}
\usepackage[hidelinks,hypertexnames=false]{hyperref}
\hypersetup{
  pdftitle={DCL-GPGLS: Dynamic Curriculum Learning for Genetic Programming Guided Local Search in Large-Scale Vehicle Routing},
  pdfauthor={Saining Liu, Yi Mei, Mengjie Zhang}
}

\newcommand{\DCL}{\textsc{DCL}}
\newcommand{\STAT}{\textsc{STAT}}
\newcommand{\DCOV}{\textsc{DCOV}}
\newcommand{\RAND}{\textsc{RAND}}
\newcommand{\CYCL}{\textsc{CYCL}}
\newcommand{\SIZE}{\textsc{SIZE}}

\begin{document}

\title{DCL-GPGLS: Dynamic Curriculum Learning for Genetic Programming Guided Local Search in Large-Scale Vehicle Routing}
\titlerunning{DCL-GPGLS}
\author{Saining Liu \and Yi Mei \and Mengjie Zhang}
\authorrunning{S. Liu et al.}
\institute{School of Engineering and Computer Science,\\
Victoria University of Wellington, Wellington, New Zealand}
\maketitle

\begin{abstract}
Genetic Programming Guided Local Search (GPGLS) uses genetic programming to evolve utility functions for guided local search in large-scale vehicle routing problems (LSVRPs). Evaluating every GP individual on every training instance at every generation is expensive, so GPGLS is usually trained on small instance batches. Existing curriculum-based GPGLS orders these batches mainly by instance size. Adaptive Curriculum Learning GPGLS (ACL-GPGLS) improves training efficiency by adapting when the search moves between fixed curriculum stages, but the instance difficulty order remains predefined. We propose DCL-GPGLS, which estimates the difficulty of each training instance from the current population's solution quality and updates the estimates during evolution. Each generation then receives a batch near a scheduled difficulty level, with a correction that limits repeated selection of the same instances. Experiments on a fixed training--test split of the CVRPLIB X set show that DCL-GPGLS achieves the best observed average rank and mean test cost among six training policies. It obtains the lowest mean cost on 36 of 65 unseen test instances and is significantly better than the static feedback-derived curriculum, matched in total evaluator calls, on 6 instances, with no significant difference on the remaining 59.
\end{abstract}


\section{Introduction}
Vehicle routing problems (VRPs) seek routes that serve geographically distributed customers while satisfying operational constraints and minimising total routing cost \cite{dantzig1959truck,toth2002vehicle,eksioglu2009vehicle}. Instances with hundreds of customers are difficult for exact optimisation, so heuristics, metaheuristics, and hyper-heuristics are commonly used to obtain high-quality solutions within practical time limits \cite{gendreau2010solving,huang2012large,arnold2019efficiently,laporte2014chapter,pecin2017improved,sabar2015math,uchoa2017new,queiroga2026xl}. Genetic Programming Guided Local Search (GPGLS) uses genetic programming (GP) to evolve utility functions that guide the penalisation decisions of guided local search (GLS) \cite{liu2024gpgls}. This reduces dependence on manually designed utility functions and allows the search guidance to be improved through evolution \cite{costa2021evolutionary}.

GPGLS training is dominated by GLS evaluation. Evaluating every GP tree on every training instance in every generation is usually impractical. A common alternative is to sample a new random subset at each generation, but the resulting fitness values can vary greatly with the selected instances. Curriculum learning (CL) provides a more structured solution: it trains on small batches and controls which instances are presented over time \cite{bengio2009curriculum,wang2021survey,soviany2022curriculum}. By replacing full-pool evaluation with scheduled batches, CL can substantially reduce the number of expensive GLS calls while preserving an informative training sequence.

Previous work introduced two curriculum mechanisms for GPGLS. CL-GPGLS uses a predefined easy-to-hard order based on customer count \cite{liu2025clgpgls}. Adaptive Curriculum Learning GPGLS (ACL-GPGLS) monitors the structural stability of the generation-best tree and the saturation of population fitness improvement to decide when to move to the next fixed curriculum stage \cite{liu2026aclgpgls}. ACL-GPGLS reduces training time by adapting stage duration, but it still uses a predefined size-based order. Customer count alone cannot distinguish instances with different demand and spatial structures, and a fixed order cannot reflect changes in the current GP population's ability \cite{smith2012measuring,gouvea2025instance}.

To address this limitation, we propose Dynamic Curriculum Learning for GPGLS (DCL-GPGLS). The method evaluates the initial population on the full training pool once to obtain an initial difficulty estimate for every instance. During evolution, the best solution produced by the current population on each selected instance provides a new difficulty observation. The estimate is updated online, and the next batch is sampled near a target difficulty that moves from easy to hard. Thus, DCL-GPGLS changes both which instances are considered difficult and which instances are selected at each generation.

The main contributions are as follows:
\begin{itemize}
    \item Extend GPGLS curriculum learning from a predefined size-based order to an online feedback-driven framework that updates instance difficulty during evolution.
    \item Develop a low-overhead difficulty estimator and batch-selection method based on population-best gaps, smoothing, a decaying initial estimate, progressive difficulty scheduling, and downweighting of frequently selected instances.
    \item Compare DCL-GPGLS with five batch-selection policies at matched main-training work, including a static feedback-derived control matched in total evaluator calls.
    \item Establish the observed aggregate performance of DCL-GPGLS on the tested split and use difficulty and selection-count diagnostics to examine, without causally isolating, possible explanations for its difference from the static control.
\end{itemize}

\section{Background}
\subsection{Genetic Programming Guided Local Search}
GPGLS is a two-level evolutionary hyper-heuristic for automatically designing GLS utility functions \cite{liu2024gpgls}. At the upper level, GP maintains a population of candidate utility functions represented as expression trees. At the lower level, each tree is incorporated into GLS and used to score candidate edges when the search reaches a local optimum. The highest-scoring edge is penalised in the augmented objective, and local search continues from the modified search landscape \cite{voudouris1999guided,arnold2019knowledge}.

Each GP tree is evaluated by running GLS on a set of training instances. The evaluator returns the best total routing cost found within a fixed time limit. These costs are converted to BKS-relative gaps and averaged over the selected batch to obtain the tree's fitness. Lower fitness is better. Elitism, tournament selection, crossover, and mutation then create the next GP population. The present work keeps the GP representation, GLS evaluator, search operators, and evolutionary settings unchanged and modifies only how the training batch is selected.

Because GLS evaluation is expensive, GPGLS uses only a subset of training instances in each generation. Randomly changing the subset reduces cost, but it also makes evolutionary selection depend on a noisy sample of instances \cite{gathercole1994dynamic,rakshit2017noisy}. The batch-selection policy is therefore an important part of GPGLS training.

\subsection{Applications of Curriculum Learning}
Curriculum learning involves two decisions: how example difficulty is defined and how examples are scheduled \cite{wang2021survey,narvekar2020curriculum,soviany2022curriculum}. Predefined curricula fix both decisions in advance. Self-paced and competence-based methods adjust the available examples to learner performance \cite{kumar2010self,platanios2019competence}, while teacher-based methods learn a sampling policy from progress or feedback \cite{graves2017automated,portelas2020teacher}.

In combinatorial optimisation, curricula have been applied across problem sizes and instance distributions \cite{lisicki2020evaluating,nabli2020curriculum,iklassov2023study}. Hardness-adaptive TSP training defines difficulty relative to the current solver \cite{zhang2022learning}, and curriculum scheduling has also been studied in online combinatorial optimisation \cite{zhou2022understanding}. These studies show that difficulty need not be a fixed property of an instance.

For GPGLS, CL-GPGLS introduced a size-based easy-to-hard sequence \cite{liu2025clgpgls}. ACL-GPGLS later adapted the duration of each curriculum stage by monitoring tree stability and population fitness improvement \cite{liu2026aclgpgls}. Both methods retain a predefined instance order. DCL-GPGLS instead updates the relative difficulty of individual instances from the current population's evaluator results and rebuilds the sampling distribution after every generation.

\section{DCL-GPGLS}

\begin{figure}[!htbp]
\centering
\includegraphics[width=0.98\linewidth]{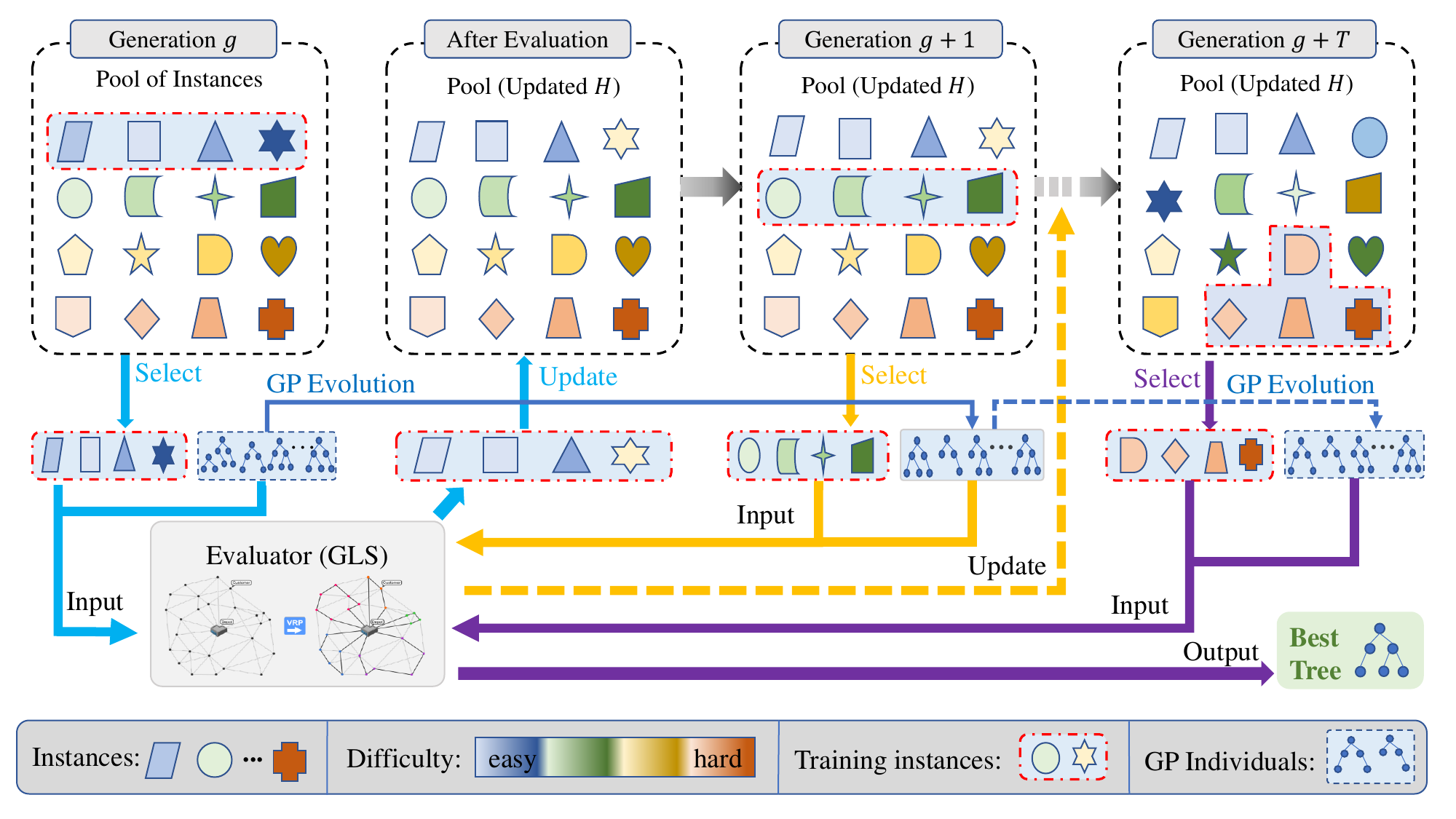}
\caption{DCL-GPGLS across generations. A selected instance batch and the current GP population are evaluated by GLS. The resulting tree fitness guides GP evolution, while the population-best gap on each selected instance updates its difficulty estimate for later batch selection.}
\label{fig:framework}
\end{figure}

Figure~\ref{fig:framework} illustrates instance selection and difficulty updating across generations.

\FloatBarrier

\subsection{Framework of DCL-GPGLS}
At generation $t$, the curriculum scheduler selects a batch $S_t$ from the training pool. Every tree in the current population $P_t$ is evaluated on the same batch. The resulting costs are used in two ways. First, the mean gap of each tree over $S_t$ gives its GP fitness. Second, the best gap achieved by the population on each selected instance gives feedback for updating that instance's difficulty estimate. Standard GP operators then generate $P_{t+1}$, while the updated difficulty estimates guide the next batch.

For tree $T$, instance $i$, and generation $t$, let $c_{T,i,t}$ denote the best total routing cost returned by GLS, and let $z_i^\star$ denote the published best-known solution (BKS). The fitness of $T$ is
\begin{equation}
F_t(T)=\frac{1}{|S_t|}\sum_{i\in S_t}
\frac{c_{T,i,t}-z_i^\star}{z_i^\star}.
\label{eq:fitness}
\end{equation}
The instance feedback used by DCL-GPGLS is the best BKS-relative gap obtained by any tree in $P_t$. This difficulty measure is specific to the current population, evaluator, and BKS values; it is not an intrinsic hardness label or a proved optimality gap.

Each training instance stores five values: an initial difficulty $d_i^0$, a smoothed difficulty $\bar d_i$, a regularised difficulty $\hat d_i$, an observation count $n_i$, and a selection count $e_i$. The value $\hat d_i$ is used for ranking and batch selection. The count $n_i$ records the number of online difficulty observations, while $e_i$ records how many times the instance has been included in a training batch after the initial full-pool evaluation. The counts have different roles but remain equal in this protocol because every selection produces one observation.

Before evolution, the initial population is evaluated once on all training instances. This evaluation initialises the difficulty values; it is not part of the warm-up batches and does not increase the selection count. The first $T_{\mathrm{warm}}$ generations then use random batches from a shuffled instance order. In the experiments, $T_{\mathrm{warm}}=5$ and $B=5$, so the warm-up contains 25 selections. It does not need to cover all 35 training instances because the full-pool initial evaluation has already provided a difficulty estimate for every instance. The warm-up only prevents the first generations from depending immediately on a single initial ranking.

\subsection{Dynamic-Difficulty Curriculum Mechanism}
\paragraph{Difficulty initialisation.}
Let $c^{\mathrm{init}}_{T,i}$ denote the cost from the one-off initial full-pool evaluation. The initial difficulty of instance $i$ is calculated from the best cost produced by the initial population:
\begin{equation}
d_i^0=
\frac{\min_{T\in P_0}c^{\mathrm{init}}_{T,i}-z_i^\star}{z_i^\star}.
\label{eq:initial}
\end{equation}
A larger value means that the initial population solves the instance less effectively. The method sets $\bar d_i=\hat d_i=d_i^0$ and $n_i=e_i=0$.

\paragraph{Online difficulty update.}
When instance $i$ is selected at generation $t$, the observed difficulty is
\begin{equation}
d_{i,t}^{\mathrm{obs}}=
\frac{\min_{T\in P_t}c_{T,i,t}-z_i^\star}{z_i^\star}.
\label{eq:observed}
\end{equation}
The smoothed and regularised estimates are updated as
\begin{align}
\bar d_i &\leftarrow (1-\alpha)\bar d_i+\alpha d_{i,t}^{\mathrm{obs}},
\nonumber\\
n_i &\leftarrow n_i+1,
\nonumber\\
\hat d_i &\leftarrow
\frac{n_i}{n_i+\kappa}\bar d_i+
\frac{\kappa}{n_i+\kappa}d_i^0,
\qquad e_i\leftarrow e_i+1.
\label{eq:update}
\end{align}
The moving average reduces short-term variation from stochastic GLS runs. The initial estimate is retained because an instance may receive only a few noisy online observations. Its weight decreases as $n_i$ increases, so well-observed instances depend mainly on current feedback. Instances not selected in generation $t$ keep their previous values.

\paragraph{Adaptive batch selection.}
Raw gaps usually decrease as the GP population improves. DCL-GPGLS therefore ranks the current values $\{\hat d_i\}$ and converts them to scores $h_i\in[0,1]$, where 0 and 1 denote the easiest and hardest current ranks. For post-warm-up generations $t=T_{\mathrm{warm}},\ldots,G-1$, the target difficulty is
\begin{equation}
\rho_t=
\left(
\frac{t-T_{\mathrm{warm}}}{G-T_{\mathrm{warm}}-1}
\right)^{\gamma}.
\label{eq:target}
\end{equation}
Thus, the target starts at 0 after warm-up and reaches 1 in the final generation. Each instance receives the unnormalised sampling weight
\begin{equation}
w_i=
\exp\!\left[-\frac{(h_i-\rho_t)^2}{2\tau^2}\right]
\frac{1}{1+e_i}.
\label{eq:weight}
\end{equation}
The first term favours instances near the current target, and $\tau$ controls the width of this difficulty band. The second term uses the selection count $e_i$ to reduce the probability of repeatedly selecting the same instances. The weights are normalised, and $B$ distinct instances are sampled without replacement.

Algorithm~\ref{alg:dcl} summarises the training procedure and returns the tree with the lowest training fitness in the final evaluated population.

\begin{algorithm}[!htbp]
\small
\caption{DCL-GPGLS training procedure}
\label{alg:dcl}
\begin{algorithmic}[1]
\Require Training pool $I$; BKS values $z_i^\star$; initial population $P_0$; generations $G$; batch size $B$; warm-up length $T_{\mathrm{warm}}$
\Ensure Final utility tree $T^\star$
\Statex \textit{Initial difficulty estimation}
\State Evaluate every $T\in P_0$ on every $i\in I$
\State Compute $d_i^0$ by Eq.~\eqref{eq:initial}; set $\bar d_i=\hat d_i=d_i^0$ and $n_i=e_i=0$
\State Generate a shuffled order of the training instances
\Statex \textit{Curriculum-guided evolution}
\For{$t=0,\ldots,G-1$}
    \If{$t<T_{\mathrm{warm}}$}
        \State Select the next $B$ instances from the shuffled order; reshuffle when exhausted
    \Else
        \State Rank $\hat d_i$ to obtain $h_i$; compute $\rho_t$ by Eq.~\eqref{eq:target}
        \State Compute $w_i$ by Eq.~\eqref{eq:weight} and sample $S_t$ without replacement
    \EndIf
    \State Evaluate every $T\in P_t$ on every $i\in S_t$ using GLS
    \State Compute $F_t(T)$ by Eq.~\eqref{eq:fitness}
    \State Compute $d_{i,t}^{\mathrm{obs}}$ by Eq.~\eqref{eq:observed} and update selected instances by Eq.~\eqref{eq:update}
    \If{$t<G-1$}
        \State Create $P_{t+1}$ by elitism, tournament selection, crossover, and mutation
    \EndIf
\EndFor
\State \Return the lowest-fitness tree in the final evaluated population as $T^\star$
\end{algorithmic}
\end{algorithm}
\FloatBarrier

Algorithm~\ref{alg:dcl} separates initial difficulty estimation, random warm-up, difficulty-based batch selection, GLS evaluation, online difficulty updating, and GP evolution. The same GLS results are used for both tree fitness and instance feedback. Therefore, after the one-off full-pool initial evaluation, DCL-GPGLS does not add extra GLS calls inside the evolutionary loop.

For $M$ training instances, ranking requires $O(M\log M)$ time per generation, weight construction requires $O(M)$, and updating the selected records requires $O(B)$. These operations are small compared with the $NB$ time-limited GLS evaluations. The only substantial additional cost is the one-off $NM$ full-pool evaluation. With $N=100$ and $M=35$, this equals 3,500 GLS evaluations and is included in the reported training budget.

\subsection{Experimental Settings}
\label{sec:exp_settings}

\paragraph{Training protocol.}
The GP representation and GLS operators follow the original GPGLS and knowledge-guided GLS configurations \cite{liu2024gpgls,arnold2019knowledge}. The main parameters are listed in Table~\ref{tab:settings}. Each policy is run independently 30 times with different random seeds. All policies perform $28\times5\times100=14{,}000$ evolutionary tree-instance evaluations. The feedback-based policies \STAT, \DCL, and \DCOV\ additionally use $35\times100=3{,}500$ initial full-pool evaluations, giving each a total budget of 17,500 evaluations. Thus, \DCL\ and \STAT\ are matched in total evaluator calls, whereas comparisons with \RAND, \CYCL, and \SIZE\ are matched only in main-training work. These call budgets do not imply equal measured wall-clock time.

\begin{table}[!htbp]
\centering
\small
\caption{Core experimental parameters. Curriculum parameters apply only to the corresponding policies.}
\label{tab:settings}
\setlength{\tabcolsep}{4.5pt}
\renewcommand{\arraystretch}{1.08}
\begin{tabularx}{\linewidth}{@{}lXl@{}}
\toprule
Component & Parameter(s) & Value(s) \\
\midrule
GP search & Population size $N$; generations $G$; batch size $B$ & 100; 28; 5 \\
Variation & Crossover; mutation; elitism probabilities & 0.80; 0.15; 0.05 \\
Representation & Maximum tree height & 6 \\
Curriculum & Warm-up $T_{\mathrm{warm}}$; band width $\tau$; exponent $\gamma$ & 5; 0.25; 1.5 \\
Difficulty update & EMA factor $\alpha$; prior strength $\kappa$ & 0.2; 5.0 \\
Evaluation & Training time per tree-instance; test time per final tree-instance & 5 s; 20 s \\
\bottomrule
\end{tabularx}
\end{table}
\FloatBarrier

\paragraph{Dataset and evaluation.}
Experiments use the 100-instance X set from CVRPLIB \cite{uchoa2017new}. In the fixed X-set ordering, the first 35 instances, X-n101 through X-n261, form the training set, and the remaining 65 instances, X-n266 through X-n1001, form the test set. These identifiers include the depot, so the corresponding customer counts are 100--260 and 265--1,000. Published BKS values are used to calculate relative gaps. The individual with the lowest training fitness in the final evaluated population is selected for testing; no validation set or held-out result is used to select this individual. This split evaluates transfer to larger unseen X-set instances, but does not isolate customer count from changes in instance identity and structure. Two-sided Wilcoxon signed-rank tests compare the final costs from 30 runs on each held-out instance using a $p$-value threshold of 0.05.

\paragraph{Compared training policies.}
Six policies use the same GPGLS framework and differ in batch selection:
\begin{itemize}
    \item \textbf{\RAND:} samples 5 distinct instances uniformly at each generation.
    \item \textbf{\CYCL:} traverses a shuffled cyclic order, so each instance is selected exactly 4 times.
    \item \textbf{\SIZE:} traverses instances in ascending order of customer count, following the size-based idea of CL-GPGLS \cite{liu2025clgpgls}.
    \item \textbf{\STAT:} estimates difficulty from feedback but fixes the difficulty order after warm-up.
    \item \textbf{\DCL:} updates difficulty throughout evolution and samples batches using Eq.~\eqref{eq:weight}.
    \item \textbf{\DCOV:} is a coverage-enhanced variant of \DCL. After warm-up, it first selects one of the least-selected instances and then samples the remaining $B-1$ instances using the DCL weights.
\end{itemize}
\STAT\ and \DCL\ use the same initial evaluation, warm-up, target schedule, downweighting of frequently selected instances, and total evaluation budget. Their only difference is whether the difficulty order is updated after warm-up, so this comparison tests the effect of online difficulty updating. \DCOV\ tests whether enforcing more uniform instance selection improves \DCL; it is not the proposed method. ACL-GPGLS is not used as a direct baseline because it changes the duration of predefined curriculum stages to reduce training time \cite{liu2026aclgpgls}, whereas the present experiment fixes all policies to the same 28-generation training process and compares how batches are constructed.

\subsection{Main Results}

\begin{figure}[!htbp]
\centering
\includegraphics[width=0.93\linewidth]{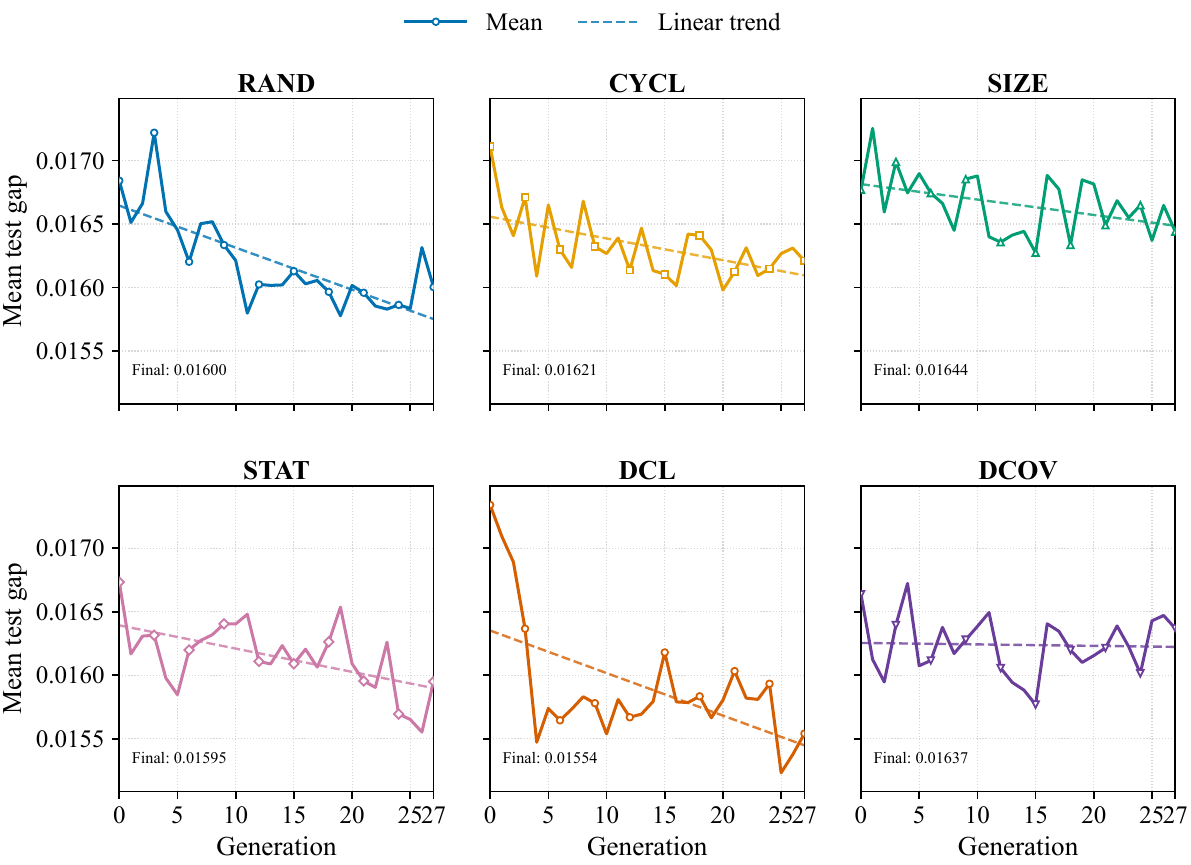}
\caption{Mean generation-wise diagnostic BKS-relative held-out gap of the six training policies. The tree with the lowest training fitness in each generation is evaluated on the held-out set, and the values are averaged over 30 independent runs. Lower values are better; held-out evaluations are used only for analysis.}
\label{fig:test_fitness}
\end{figure}

\paragraph{Generation-wise performance.}
Figure~\ref{fig:test_fitness} shows the mean BKS-relative held-out gap of the six training policies over 28 generations and 30 independent runs. Each generation's training-best tree is evaluated retrospectively on all 65 held-out instances. These diagnostic evaluations do not feed back into GP training, parent selection, or final-tree selection and are separate from the training budgets above.

All policies improve during evolution, but their improvement rates differ. \CYCL, \SIZE, and \DCOV\ decrease relatively slowly, whereas \RAND, \STAT, and \DCL\ improve more rapidly during the early generations. \DCL\ reaches a low-fitness region earlier and achieves the lowest mean diagnostic gap in 20 of the 28 generations. At the final generation, \DCL\ obtains the lowest mean diagnostic gap of 0.01554, followed by \STAT\ with 0.01595, while the remaining policies range from 0.01600 to 0.01644. These descriptive BKS-normalised trajectories provide temporal context; they are distinct from the final raw-cost and instance-wise rank summaries below, which weight instances differently.

\paragraph{Final performance.}
Table~\ref{tab:main} reports the aggregate final-test results and the unadjusted instance-wise Wilcoxon signed-rank decisions.

\begin{table}[!htbp]
\centering
\footnotesize
\caption{Final test performance on 65 instances. Avg. rank and Rank SD denote the mean and standard deviation of the per-instance method ranks. \#Lowest denotes the number of instances with the lowest mean cost. The last three columns report the unadjusted per-instance Wilcoxon signed-rank results for \DCL\ relative to each comparator.}
\label{tab:main}
\setlength{\tabcolsep}{2.15pt}
\renewcommand{\arraystretch}{1.05}
\begin{tabular}{@{}lrrrrrrr@{}}
\toprule
& \multicolumn{4}{c}{Aggregate performance}
& \multicolumn{3}{c}{\DCL\ relative to comparator} \\
\cmidrule(lr){2-5}\cmidrule(l){6-8}
Method & Avg. rank & Rank SD & \#Lowest & Mean cost
& Sig. better & No sig. diff. & Sig. worse \\
\midrule
\DCL  & 1.83 & 1.24 & 36 & 81590.64 & -- & -- & -- \\
\STAT & 3.32 & 1.50 &  9 & 81638.14 &  6 & 59 & 0 \\
\RAND & 3.31 & 1.51 &  7 & 81644.66 & 12 & 53 & 0 \\
\CYCL & 3.49 & 1.81 & 10 & 81637.16 & 17 & 48 & 0 \\
\SIZE & 4.49 & 1.20 &  0 & 81674.03 & 24 & 41 & 0 \\
\DCOV & 4.55 & 1.39 &  3 & 81661.49 & 14 & 51 & 0 \\
\bottomrule
\end{tabular}
\end{table}
\FloatBarrier

The left part of Table~\ref{tab:main} reports the aggregate results. \DCL\ achieves the lowest average rank of 1.83 and the lowest mean cost of 81590.64. It also obtains the lowest mean cost on 36 of the 65 test instances. The other policies have average ranks between 3.31 and 4.55 and obtain the lowest mean cost on at most 10 instances.

The right part of Table~\ref{tab:main} reports the per-instance Wilcoxon test results. \DCL\ is significantly better than \STAT, \RAND, \CYCL, \SIZE, and \DCOV\ on 6, 12, 17, 24, and 14 instances, respectively. It shows no significant difference on the remaining 59, 53, 48, 41, and 51 instances. \DCL\ is not significantly worse than any policy on any test instance. Overall, \DCL\ has the best observed aggregate performance among the six training policies under this protocol; these results do not establish superiority on every instance.

\begin{figure}[!htbp]
\centering
\includegraphics[width=0.98\linewidth]{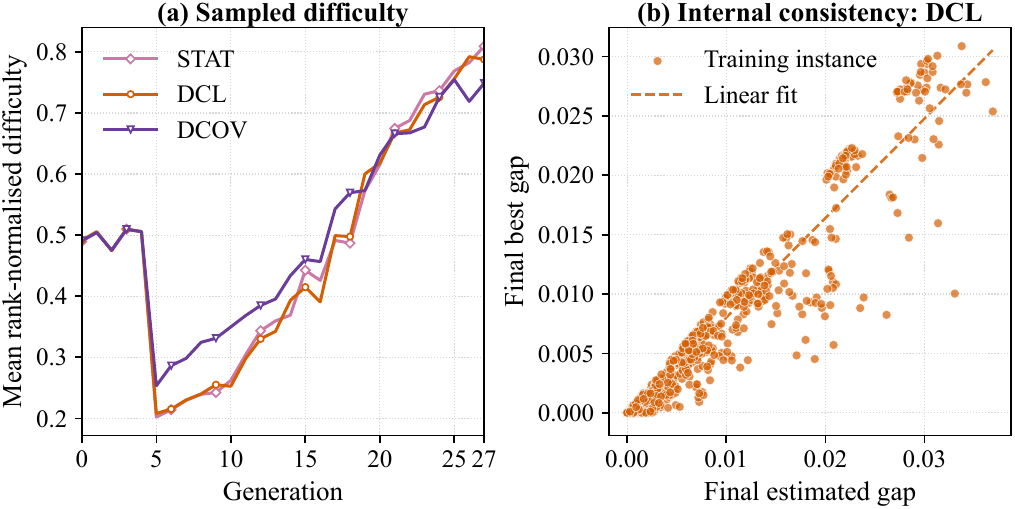}
\caption{Mechanism analysis: (a) mean sampled rank-normalised difficulty over generations; (b) final regularised difficulty estimate against the latest stored population-best observed gap for \DCL, pooled over instance--run records.}
\label{fig:mechanism}
\end{figure}

\paragraph{Mechanism analysis.}
Figure~\ref{fig:mechanism} examines sampled difficulty and the consistency of the stored difficulty estimates.

\FloatBarrier

Figure~\ref{fig:mechanism}(a) shows the mean rank-normalised difficulty of the instances selected by the feedback-based policies. After warm-up, all three policies gradually sample harder difficulty bands. \STAT\ and \DCL\ follow similar trajectories under their shared target schedule and downweighting of frequently selected instances. This weakens an explanation based solely on one policy sampling globally easier or harder batches. \STAT\ fixes the difficulty order after warm-up, whereas \DCL\ updates it during evolution. They may therefore select different instances even when their batches have similar mean difficulty.

Figure~\ref{fig:mechanism}(b) compares the final regularised difficulty estimate with the latest stored population-best observed gap for each instance--run record under \DCL. An unsampled instance retains its earlier observation, so this gap need not come from the final GP population. Instances with higher estimated difficulty generally retain larger observed gaps. Because both quantities are obtained from related evaluator feedback, the figure is an internal consistency check rather than an independent validation of instance hardness.

Each training instance is selected four times on average. For both \STAT\ and \DCL, the selection counts have a standard deviation of 1.24 and a range of 1--8 after rounding. \DCOV\ selects instances more uniformly but has worse aggregate performance. These summaries do not support more uniform aggregate exposure as the explanation of the \DCL--\STAT\ difference, but they do not establish identical count vectors or isolate revisit timing. The diagnostics are consistent with a benefit from dynamic scheduling, rather than a causal demonstration of that mechanism.

\section{Conclusions}
This paper extends GPGLS from a predefined size-based curriculum to an online feedback-driven curriculum. DCL-GPGLS estimates instance difficulty from the current population's evaluator results and updates the estimates during evolution. It combines initial full-pool estimation, smoothed online updates, a decaying initial estimate, progressive difficulty scheduling, and downweighting of frequently selected instances to construct each training batch.

The results show that DCL-GPGLS achieves the best observed average rank and mean cost among the six training policies and obtains the lowest mean cost on 36 of the 65 unseen test instances. Compared with \STAT, matched in total evaluator calls, \DCL\ is significantly better on 6 instances and shows no significant difference on 59, with no significantly worse result. The mechanism analysis narrows possible explanations but does not isolate the causal effect of revisit timing.

These findings are specific to one fixed X-set split, one parameter configuration, and the evaluated time limits. Difficulty estimates can become stale when an instance is not selected. Moreover, \RAND, \CYCL, and \SIZE\ do not incur the extra full-pool initialisation cost, so their comparisons with \DCL\ are not matched in total training calls. The descriptive diagnostics have no uncertainty intervals and do not independently validate a universal instance-hardness measure.

Future work will examine alternative data splits and parameter settings, incorporate additional difficulty signals, reduce the initial full-pool evaluation cost, and apply DCL-GPGLS to other routing variants and evolutionary optimisation frameworks.

\begingroup
\interlinepenalty=10000
\bibliographystyle{splncs04}
\bibliography{ref}
\endgroup
\end{document}